\pdfoutput=1

\documentclass[11pt]{article}

\usepackage[]{EMNLP2023}

\usepackage{times}
\usepackage{latexsym}
\usepackage{amsfonts}
\usepackage{amsmath}
\usepackage{makecell}
\usepackage{svg}
\usepackage{bbm}
\usepackage{booktabs}
\usepackage{tikz}
\usetikzlibrary{arrows.meta,
                backgrounds,
                fit,
                matrix}
                
\usepackage[inline]{enumitem}

\DeclareMathOperator*{\argmax}{arg\,max}
\newcommand{\sep}{\langle \textsc{SEP} \rangle}
\def\doubleunderline#1{\underline{\underline{#1}}}

\usepackage[T1]{fontenc}
\usepackage[utf8]{inputenc}

\usepackage{microtype}
\usepackage{xspace}

\newcommand{\mucow}{\textsc{MuCoW}\xspace}
\newcommand{\docmucow}{\textsc{doc-MuCoW}\xspace}

\newcommand{\tfidf}[1]{\texttt{tfidf}$_{#1}$}
\newcommand{\tfidfshuf}[1]{\texttt{tfidf}$_{#1}^{\text{shuf}}$}
\newcommand{\yake}[1]{\texttt{yake}$_{#1}$}
\newcommand{\yakeshuf}[1]{\texttt{yake}$_{#1}^{\text{shuf}}$}
\newcommand{\doc}{\texttt{2sent}\xspace}
\newcommand{\sent}{\texttt{sent}\xspace}

\newcommand{\bleu}{\textsc{BLEU}\xspace}
\newcommand{\comet}{\textsc{COMET}\xspace}
\newcommand{\cprec}{\textsc{P}\xspace}
\newcommand{\crec}{\textsc{R}\xspace}
\newcommand{\cf}{\textsc{F1}\xspace}
\newcommand{\mt}{MT\xspace}

\newcommand{\wsd}{WSD\xspace}

\newcommand{\len}{Training time\xspace}

\title{Improving Word Sense Disambiguation \\ in Neural Machine Translation with Salient Document Context}

\author{Elijah Rippeth \and Marine Carpuat \\
  Department of Computer Science \\
  University of Maryland \\
  \texttt{\{erip,marine\}@cs.umd.edu} \\\And
  Kevin Duh \and Matt Post \\
  HLTCOE \\
  Johns Hopkins University \\
  \texttt{\{kevinduh,post\}@cs.jhu.edu} \\}

\begin{document}
\maketitle
\begin{abstract}
Lexical ambiguity is a challenging and pervasive problem in machine translation (\mt). We introduce a simple and scalable approach to resolve translation ambiguity by incorporating a small amount of extra-sentential context in neural \mt. Our approach requires no sense annotation and no change to standard model architectures. Since actual document context is not available for the vast majority of \mt training data, we collect related sentences for each input to construct pseudo-documents. Salient words from pseudo-documents are then encoded as a prefix to each source sentence to condition the generation of the translation. To evaluate, we release \docmucow, a challenge set for translation disambiguation based on the English-German \mucow \cite{raganato-etal-2020-evaluation} augmented with document IDs. Extensive experiments show that our method translates ambiguous source words better than strong sentence-level baselines and comparable document-level baselines while reducing training costs.
\end{abstract}

\section{Introduction}

Lexical ambiguity is a challenging problem for machine translation (\mt) systems---when an ambiguous source word lacks sufficient context, a translation system may fail to translate it accurately, as seen in Table \ref{ex:example}. Exacerbating the situation, neural \mt systems are nearly always trained, evaluated, and deployed at the sentence level by translating individual sentences independently, ignoring context when available. As a result, sentence-level systems typically struggle with word sense disambiguation (\wsd) \cite{rios-gonzales-etal-2017-improving} where extra-sentential context could help \cite{bawden-etal-2018-evaluating}. 

Document-level \mt has attempted to address this issue by incorporating additional sentence context. While interest in document-level \mt has increased recently \cite[i.a.]{rysova-etal-2019-test, farajian-etal-2020-findings, akhbardeh-etal-2021-findings}, its adoption has been slow in part due to architectural complexity, computational expense, and lack of high-quality document-level data \cite{sun-etal-2022-rethinking}. 
  
\begin{table}[t]
\centering
\scalebox{0.8}{
\begin{tabular}{ll} \toprule
English src         & I went to the \textbf{bank} after lunch.                           \\ 
\midrule
context$_1$     & finance money vault checking deposit                                     \\ 
German ref$_1$         & \makecell[l]{Ich ging zur {\color{blue}Bank} nach dem Mittagessen.}             \\ 
\midrule
context$_2$     & nature river water forest shore                                    \\ 
German ref$_2$         & \makecell[l]{Ich ging zum {\color{red}{Ufer}} nach dem Mittagessen.} \\ \bottomrule
\end{tabular}}
\caption{Different contexts give rise to different German translations of the ambiguous English word \textbf{bank}: a {\color{blue} financial institution} or a {\color{red} shore}.}
\label{ex:example}
\end{table}

To address these limitations, we produce context by prefixing the source sentence with a document ``summary''---this summary is a bag of salient words which contextualizes the sentence in a noisily compiled pseudo-document. We hypothesize this approach can improve accuracy of ambiguous word translation at the sentence-level with no additional parameters. 

To this end, we contribute the following:
\begin{enumerate}[label=\textbf{C\arabic*}]
    \item A framework for incorporating salient context into training, which reduces context to relevant keywords while maintaining a standard sentence-level architecture for \mt.
    \item An updated test set, English-German \docmucow, which is a document ID-augmented subset of \mucow \cite{raganato-etal-2020-evaluation}: a challenging test set to measure performance of systems in the task of \wsd in \mt. 
    \item A set of results and accompanying analysis showing that saliency-based models uniformly improve over sentence-level and comparable document-level baselines in translation disambiguation at a reduced training cost.
\end{enumerate}

\section{Method}\label{method}

\paragraph{Pseudo-Documents} Because \mt is primarily framed as a sentence-level task, the vast majority of corpora do not preserve document information, hindering document-level approaches to \mt \cite{li-etal-2020-multi-encoder,sun-etal-2022-rethinking}. We address this problem by relaxing the requirement for actual document context. Instead, we define ``pseudo-documents'' which are bags of related sentences constructed in a potentially fuzzy manner. The related sentences could be drawn from the same actual document or from webpages with similar URLs, they could belong to the same topic model cluster, etc. While this is a noisy process that does not provide the structure and ordering of actual documents, the resulting pseudo-documents are still likely to provide most context information relevant to \wsd such as topical keywords.

\paragraph{Saliency-based \mt models}

We introduce a framework for incorporating salient document context into training through keyword prefixing. We define a sentence $s_i := \left[ w_1, \dots, w_n \right] \in \mathbb{S} \subset \mathbb{W}^n$ as a sequence of $n$ words and a pseudo-document $d_j := \{ s_1, \dots, s_m \} \in \mathbb{D} \subset \mathbb{S}^m$ as a set of $m > 1$ sentences. Given a pseudo-document $d \in \mathbb{D}$, and a saliency function $\phi_n: \mathbb{D} \rightarrow \mathbb{W}^n$ which maps a pseudo-document to a set of $n$ words, we define the $k$ most salient words given $d$ as $\hat{\textbf{w}}^k = \phi_k(d)$\footnote{in descending order by saliency function-specific weight.\label{ordering_footnote}}.

For a given source sentence $x \in d_j$ and salient words given $x$'s parent pseudo-document, $\hat{\textbf{w}}^k = \phi_k(d_j)$, we generate a translation according to $$y^* = \argmax p(y \mid x, \hat{\textbf{w}}^k)$$ allowing the model to condition on a \textit{small amount} of extra-sentential information from $x$'s ``global context''. To do this, we treat $\hat{\textbf{w}}^k$ as a prefix of a modified model input as given by $$x' = \hat{\textbf{w}}^k \oplus \left[ \sep \right] \oplus x$$ where $\oplus$ is defined as sequence concatenation and $x'$ is the new effective source sentence. This process is applied to both training and test samples. The target side of the training data remains unchanged. A benefit of this simple source sentence manipulation is that by treating $x'$ as the source sentence, we maintain an identical model architecture, loss function, and decoding process as traditional sentence-level neural \mt systems.

An example of these salient words and their impact can be seen in Table \ref{ex:example} in which a given source sentence $x$ being prefixed with varying salient context $\hat{\textbf{w}}^{5}$ yields different, valid translations with varying realizations of the ambiguous source word. 

\paragraph{Saliency functions} We consider two saliency functions for extracting representative words from each pseudo-document. 

First, we use the standard term frequency inverse document frequency (\textbf{tfidf}) score \cite{sparckjones_statistical_1972}, which views as salient words that are frequent in a corpus $D$ but occur in relatively few documents. Formally, for a document $d$,
\begin{align*}
    \phi^{tfidf}_n(d) = \argmax_{\hat{\textbf{w}}^n \subset d} \ \{ \operatorname{tfidf}_D(w, d) : w \in \hat{\textbf{w}}^n \}
\end{align*}
with
\begin{align*}
    \operatorname{tfidf}_D(w, d) & = \operatorname{tf}(t, d) \cdot \operatorname{idf}_D(t) \\
    \operatorname{tf}(t, d) & = \frac{1}{|d|}\sum_{t' \in d} \mathbbm{1}[t = t'] \\
    \operatorname{idf}_D(t) & = \log\left(\frac{|D| + 1}{\operatorname{df}_D(t) + 1}\right) + 1 \\
    \operatorname{df}_D(t) & = \sum_{d \in D} \mathbbm{1}[t \in d]
\end{align*}

Second, we use \textbf{YAKE!} \cite{CAMPOS2020257}, a strong off-the-shelf unsupervised keyword extractor. Unlike tfidf, YAKE! does not treat documents as bags-of-words, and uses features such as candidate position within a document, candidate term frequency, and duplicity of candidates across sentences. YAKE! was found to be likelier to return highly-ranked human-generated keywords when compared to contemporary state-of-the-art unsupervised statistical, unsupervised graph-based, and supervised baselines across a number of diverse corpora and across a number of languages \cite{CAMPOS2020257}.  By fixing hyperparameters, we can treat YAKE! as a black-box keyword extractor which returns the $n$ highest scoring words in a document with the following saliency function:
\begin{align*}
    \phi^{yake}_n(d) = \underset{topk=n}{\operatorname{YAKE}}(d)
\end{align*}

\section{\docmucow: A Large Test Set for \wsd with Document Context}

\begin{table*}[t]
\centering
\begin{tabular}{lrrr} \toprule
    & \# ambiguous  types & \# sentences & \# doc IDs \\ \midrule
\multicolumn{4}{l}{\textit{Subcorpus}}\\
~~Europarl v7 \cite{koehn-2005-europarl}     & 199                  & 8912        & 871         \\
~~GlobalVoices v2017q3 \cite{tiedemann-2012-parallel} & 160                  & 1395        & 456         \\
~~JW300 \cite{agic-vulic-2019-jw300}           & 204                  & 6505        & 1944        \\
~~MultiUN v1 \cite{eisele-chen-2010-multiun}      & 14                   & 25          & 11          \\
~~TED 2013 v1.1 \cite{cettolo-etal-2013-report}          & 194                  & 3708        & 667        \\
\textit{Combined}          & 206                  & 20545       & 3949      \\ \bottomrule
\end{tabular}
\caption{The composition of the \docmucow test set shows that is has broad document coverage and a wide range of ambiguous word types. Due to its large size, smaller improvements are more meaningful.}
\label{tab:docmucow}
\end{table*}

Standard \mt test sets are designed to measure translation quality and do not afford a direct measure of translation disambiguation accuracy.  In past work, targeted \wsd evaluation has used challenge sets \citet{rios-gonzales-etal-2017-improving}, where systems are evaluated by scoring correct and incorrect translations of a source sentence constructed to feature lexical ambiguity. While enabling controlled comparisons, this approach is limited to evaluating the scores of \mt systems on artificially constructed data, rather than their generation ability on naturally occurring text.
The \mucow dataset bridges that gap by automatically augmenting standard \mt test sets from diverse domains with sense annotation \cite{raganato-etal-2020-evaluation}. Each example in \mucow contains an ambiguous source lemma, the ambiguous token's sense cluster ID, the corpus from which the bitext was sourced, the bitext, and the ambiguous source token. \mucow derives its sense inventory from  BabelNet \cite{navigli-ponzetto-2010-babelnet}: BabelNet-derived embeddings \cite{mancini-etal-2017-embedding} are used to form sense clusters for translations of ambiguous words occurring in \mt test sets.

A key missing attribute of each example for the purposes of this work is the document ID. Without these document IDs we cannot create document context with which to test our method. To this end, we reconstruct document IDs for examples drawn from available sources\footnote{\textbf{Tatoeba}, the \underline{Books Corpus}, \underline{CommonCrawl}, and the \underline{EU Bookshop Corpus} are either \textbf{sentence-level in nature} or \underline{unavailable in a form permitting doc ID assignment}.} by searching the raw text files from the original corpora for sentences in \mucow \footnote{To address differences in tokenization, we remove all non-alphanumeric characters from the needle and the haystack and check for substring inclusion}. The resulting test set, \docmucow, contains 20.5k examples from \mucow augmented with document IDs, details of which can be found in Table \ref{tab:docmucow}.
\paragraph{\wsd evaluation}\label{mucow_eval}

Along with the challenge sets, \citet{raganato-etal-2020-evaluation} prescribe an evaluation scheme to measure the corpus-level precision (\cprec), recall (\crec), and \cf of target senses in system outputs. To measure this, system outputs are lemmatized using an off-the-shelf parsing model and sense inventories are queried relative to the source sentence's ambiguous lemma. Formally, given system output $s$ as a set of lemmas, a set of acceptable ``positive'' lemmatized realizations, $p$, for the source sentence's ambiguous token, and a set of unacceptable ``negative'' lemmatized realizations, $n$, for the source sentence's ambiguous token, the sense accuracy labels are computed as follows:
\[
\operatorname{C}(s, p, n) = 
    \begin{cases}
    \textsc{pos} & s \cap p \neq \emptyset \land s \cap n = \emptyset \\
    \textsc{neg} & s \cap n \neq \emptyset \\
    \textsc{unk} & \text{otherwise}
  \end{cases}
\]
with
\begin{align*}
\text{\cprec} & = \frac{\# \textsc{pos}}{\# \textsc{pos} + \# \textsc{neg}} \text{ and} \\
\text{\crec} & = \frac{\# \textsc{pos}}{\# \textsc{pos} + \# \textsc{neg}  + \# \textsc{unk}}
\end{align*}
While \crec is a slightly unusual definition of recall due to lack of distinction between true and false negatives afforded by $\operatorname{C}$, we use it for consistency. 

\section{Experimental Setup}\label{experimental_setup}

\subsection{Training Data}

\citet{sun-etal-2022-rethinking} compile a comprehensive list of English-German document-level datasets which shows there are only on the order of 2M sentences (across approximately 125k documents) of document-level data. A subset of these (e.g., Europarl) are contained within \docmucow, making them unavailable for training. Additionally, the nature of documents in several of these datasets (e.g., OpenSubtitles, Europarl) would make the documents too large and thus impractical for providing salient context for translation disambiguation.

Instead, we can construct pseudo-documents for training data using ParaCrawl \cite{banon-etal-2020-paracrawl}. To do this, we process the English-German TMX representation of filtered ParaCrawl\footnote{We retain pairs of sentences which have cosine similarity greater than $0.85$ based on \texttt{LaBSE} representations \cite{feng-etal-2022-language}.} which provides a set of source and target URLs, treating canonicalized versions of sets of URLs as proxies for document IDs. These are pseudo-documents, rather than real documents, since the canonicalization process is noisy, and since the resulting documents are bags-of-sentences. We filter long documents, keeping only documents which contain between $2$ and $10$ sentences. This results in 9,412,822 pseudo-documents containing a total of 47,418,165 high-quality bitext pairs.

\subsection{Systems}

\paragraph{Baselines} As a baseline we consider two systems: context-agnostic \sent and context-aware \doc. \sent is trained on the sentences from all pseudo-documents in the training set. \doc is trained with all sentences from all pseudo-documents in the training set, but each source sentence $x \in d_j$ is prefixed with a randomly selected sentence from the set of sentences within $x$'s parent pseudo-document, omitting $x$; i.e., $\hat{x} \in d_j - \{x\}$. We train a joint unigram LM segmenter \cite{kudo-2018-subword} using SentencePiece \cite{kudo-richardson-2018-sentencepiece} by sampling 10m sentences from the training data with 0.995 character coverage, a vocab size of 32k, and a user defined $\langle \text{SEP} \rangle$ token which is only used in context-aware models. 

We train systems $\texttt{X}_N$ using the same architecture and training data as \sent, but with context constructed by passing pseudo-documents to saliency function $\phi^{X}_{N}$ as descibed in Section \ref{method}, with $N$ chosen to be $5$ and $10$. This results in four systems: \tfidf{5}, \tfidf{10}, \yake{5}, and \yake{10}.

Each system follows an identical architecture: a 60.5m parameter $6$+$6$ transformer \cite{NIPS2017_3f5ee243} trained to optimize label smoothed cross-entropy using Adam, trained with the fairseq toolkit \cite{ott-etal-2019-fairseq}. We split the training data into four shards with equal number of examples and train for $30$ epochs\footnote{an epoch is defined as a pass over a shard} with a maximum of $32,768$ tokens per batch on a single A6000 GPU.

\paragraph{Saliency-based models} For salient context, pseudo-documents are lowercased and tokenized with Moses. A single tfidf statistic was learned from a sample of 10m pseudo-documents in the training data and used for all \tfidf{} systems. \yake{} systems were presented with documents with identical preprocessing as the tfidf based systems. For all \yake{} systems we use unigram features to ensure we only select key \textit{words} (cf.\ phrases) and use the ``sequence matcher'' keyword deduplication algorithm with a Levenshtein ratio of $0.9$.

\begin{table*}[t]
\centering
\scalebox{0.9}{
\begin{tabular}{lcccrrrr}
\toprule
& \multicolumn{3}{c}{\wsd Metrics} & \multicolumn{2}{c}{Translation Quality} & \multicolumn{2}{c}{Efficiency} \\ 
System          & \cprec  & \crec & \cf & \bleu  & \comet & Length & \len \\ \midrule
\multicolumn{7}{l}{\textit{Baseline models}}\\
\sent      & 0.7850   & 0.6008   & 0.6807    & 21.8  & 0.782 & 24.3 & 97.8 ($\times 1.00$) \\ 
\doc      & 0.7830   & 0.5976   & 0.6779    & \doubleunderline{21.9}  & 0.783 & 50.5 & 135.4 ($\times 1.38$)\\ 
\multicolumn{7}{l}{\textit{Saliency-based models}}\\
\tfidf{5}          & 0.7816 &	0.5954 &	0.6759    & \doubleunderline{21.9}  & \doubleunderline{0.784} & 36.0 & 113.3 ($\times 1.16$) \\
\tfidf{10}         & \underline{0.7871} &	\underline{0.6013} &	\underline{0.6817}    & \doubleunderline{22.0}  & 0.783 & 44.6 & 127.5 ($\times 1.30$)\\ 
\yake{5}          & \underline{0.7878} &	0.5973 &	0.6795     & \doubleunderline{22.0}   & \doubleunderline{0.785} & 34.3 & 110.7 ($\times 1.13$) \\
\yake{10}          & \underline{0.7885} &	\underline{0.6058} &	\underline{0.6852}     & 21.9   & 0.783 & 41.7 & 122.8 ($\times 1.26$) \\ 
\bottomrule
\end{tabular}}
\caption{Significant improvements between systems and \sent at the 95\% CI as determined by \underline{one-sided t-test} with 50 trials each with 750 samples (with replacement) and as determined by  \doubleunderline{paired bootstrap resampling} with 5k resamples. Saliency-based systems perform comparably or better than \sent and \doc in \wsd metrics and Translation Quality, but at a fraction of the training time and frequently with significant improvements. Because of the size of \docmucow (20.5k sentences), large changes are unnecessary for statistical significance.}
\label{tab:main_results}
\end{table*}

\subsection{Evaluation}

Our evaluation targets three main goals: ambiguous word translation accuracy, general \mt quality, and training time.

To address the first goal, we use the \mucow evaluation pipeline described in Section \ref{mucow_eval} and record \crec, \cprec, and \cf. We use use spaCy v3.6.1 and \texttt{de\_core\_news\_lg} v3.6.0 \cite{ines_montani_2023_8225292} to lemmatize system outputs.

To address the second goal, we measure \bleu \cite{papineni-etal-2002-bleu} using SacreBLEU\footnote{\tiny{\texttt{nrefs:1|case:mixed|eff:no|tok:13a|smooth:exp|version:2.3.1}}} \cite{post-2018-call} and \comet \footnote{\tiny{\texttt{Unbabel/wmt22-comet-da}}} \cite{rei-etal-2020-comet}. We expect neither of these to vary dramatically and include them to ensure no regressions in translation quality. We expect that the change in translation of individual words would cause \bleu to improve only slightly. The expectations for \comet are less clear: if additional context is required, \comet scores \textit{may} vary widely since the semantics of the translations could change dramatically; otherwise context-aware models may perform as well as context-agnostic leading to small \comet score differences.

To address the third goal, we also record the average sequence length in number of subwords and training time in thousands of seconds, computing the increase in training time relative to \sent. As attention costs grow quadratically in sequence length, maintaining a tradeoff between task performance and training time is critical for practical purposes and is a benefit of using \textit{few} salient words.

\section{Main Results}

\paragraph{\wsd metrics} As reported in Table \ref{tab:main_results}, most saliency-based systems significantly improve ambiguous word translation relative to the sentence-only baseline \sent in some respect. Comparatively, the document-level system \doc actually perform worse than \sent. Systems trained with fewer salient words ($5$) translate ambiguous words worse than systems trained with $10$ salient words, regardless of the saliency function used. The saliency-based model with a more sophisticated saliency function, \yake{}, yields better \wsd results than \tfidf{}. 

\paragraph{Translation quality} As expected, extra-sentential context preserves or slightly improves the translation quality of the sentence-level model, based on \bleu and \comet scores. Saliency-based models and \doc models achieve comparable translation quality overall. The saliency-based \tfidf{5} and \yake{5} models stands out by significantly improving \bleu and \comet over the \sent baseline.

While \doc has significantly different \bleu from \sent, inspection of the sentence-level \bleu scores suggests that context helps the system match the reference style, even if it does not produce the correct sense.

\paragraph{Efficiency} The benefits of extra-sentential context come at a much smaller training cost when using saliency-based models than document-level models. Because \doc's are the longest inputs of all models in this work, its training time is considerably longer. Saliency-based models train 8-25\% faster than \doc with identical architectures, hyperparameters, and hardware while performing comparably or better in other aspects of our evaluation.

Overall, our results show that, in the absence of actual document context, pseudo-documents provide useful context to translate ambiguous words better. Further, summarizing pseudo-documents using a small number of salient words is an effective strategy to improve translation of ambiguous words, while reducing training time costs compared to document-based systems.  Even with these advances, there remains room for improvement in \wsd metrics on \docmucow, suggesting that \wsd is not a solved problem in high-resource \mt.

\section{Analysis}

Next, we conduct an extensive analysis to better understand these results, starting by manually inspecting outputs, before breaking down quantitative results across several data and modeling dimensions.
  
\subsection{Manual Inspection} 

We manually inspect $50$ random samples of examples where \wsd results differ across models.

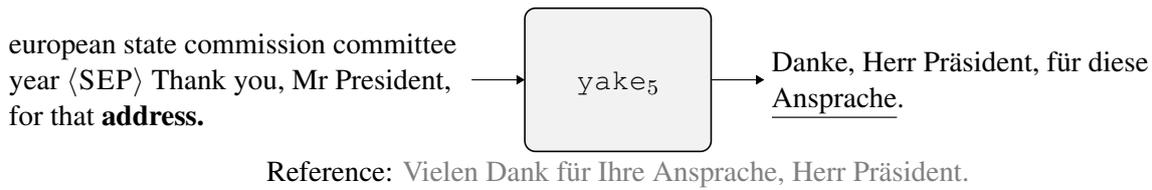
\begin{figure*}[t]
    \centering
\begin{tikzpicture}[scale=0.8]
    \matrix (m) [
        matrix of nodes,
        nodes in empty cells,
        nodes = {
            text height=2ex, text depth=0.5ex,
            inner sep=1mm, outer sep=0mm,
        },
        row sep = 0mm, column sep = 7mm,
        column 1/.style = {nodes={align=right, anchor=south east}},
        column 2/.style = {nodes={align=center, anchor=south, text width=13ex}}, %
        column 3/.style = {nodes={align=left, anchor=south west}},
    ] {
                 &                   &       \\
       \makecell[l]{european state commission committee \\ year $\sep$ Thank you, Mr President, \\ for that \textbf{address.} } &   \yake{5}   &   \makecell[l]{Danke, Herr Präsident, für diese \\ \underline{Ansprache}.}                \\
       &                   &                  \\
    };
    \scoped[on background layer]
        \node (enc)  [draw, rounded corners, semithick, fill=gray!10,
                      inner sep = 0mm, outer sep= 0mm,
                      fit=(m-1-2) (m-3-2)] {};
    \foreach    \i in {2}
        \draw[-{Triangle[angle=60:2pt 4]}]    (m-\i-1) -- (m-\i-2);
    \foreach    \i in {2}
        \draw[-{Triangle[angle=60:2pt 4]}]    (m-\i-2) -- (m-\i-3);
      \draw (m-3-2.south) coordinate node[below, node distance = 4em]{Reference: {\color{gray}Vielen Dank für Ihre \underline{Ansprache}, Herr Präsident.}} (m.south east);

\end{tikzpicture}
    \caption{Here context appropriately contextualizes the \textbf{ambiguous word}, nudging the model toward the sense of \underline{``speech''}}\vspace{0.5cm}%
    \label{ex:pos_examples}
\end{figure*}

When saliency-based methods improve over the \sent baseline, the extracted salient words often provide topical information needed to select the appropriate sense, as seen in Figure \ref{ex:pos_examples}.  When saliency-based methods are worse than the \sent baseline, the salient context appears to be useful for summarizing the document, but is not useful for disambiguation of the sentence being translated. This suggests that correctly selecting context is important and the lack of local context may hurt translation disambiguation in these cases. Additionally we find that in very rare cases, noise from the automatic curation of \mucow penalizes correct generations by assigning synonyms different sense cluster IDs. We present full examples in Figures \ref{ex:orthogonal}--\ref{ex:mucow_noise} in the Appendix.

\begin{figure*}[t!]
    \centering
\begin{tikzpicture}[scale=0.8]
    \matrix (m) [
        matrix of nodes,
        nodes in empty cells,
        nodes = {
            text height=2ex, text depth=0.5ex,
            inner sep=4mm, outer sep=0mm,
        },
        row sep = 0mm, column sep = 7mm,
        column 1/.style = {nodes={align=right, anchor=south east}},
        column 2/.style = {nodes={align=center, anchor=south, text width=13ex}}, %
        column 3/.style = {nodes={align=left, anchor=south west}},
    ] {
                 &                   &       \\
       \makecell[l]{To put this into context, more people \\ will die in the European Union from \\ the effects of a strain of TB that is \\ drug - resistant than will be affected \\ by avian flu. $\sep$ Why not pick an \\ easy \textbf{target}, one that does not \\ move about?} &   \doc   &   \makecell[l]{Warum nicht ein einfaches \underline{Ziel} \\ wählen, das sich nicht \\ bewegt<unk>}                \\
       &                   &                  \\
    };
    \scoped[on background layer]
        \node (enc)  [draw, rounded corners, semithick, fill=gray!10,
                      inner sep = 0mm, outer sep= 0mm,
                      fit=(m-1-2) (m-3-2)] {};
    \foreach    \i in {2}
        \draw[-{Triangle[angle=60:2pt 4]}]    (m-\i-1) -- (m-\i-2);
    \foreach    \i in {2}
        \draw[-{Triangle[angle=60:2pt 4]}]    (m-\i-2) -- (m-\i-3);
      \draw (m-3-2.south) coordinate node[below, node distance = 4em]{Reference: {\color{gray}Warum sucht man sich eigentlich kein leichteres Opfer, eines, das sich nicht bewegt?}} (m.south east);
\end{tikzpicture}
    \caption{Here context appropriately contextualizes the \textbf{ambiguous word}, but this cue is not strong enough to appropriately translate it. sys realization is the \underline{``goal''} sense of target rather than ``victim''. We note that the inclusion of <unk> in the output is an unfortunate effect of randomness in the subword model training.}
    \label{ex:randsent_example}
\end{figure*}
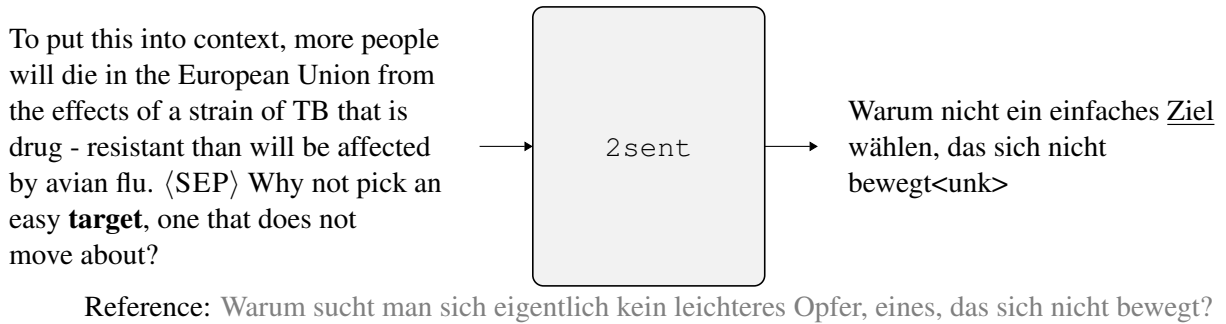

Among cases where \doc made an incorrect sense prediction, the input often contained sufficient support for the correct decision, but  \doc failed to incorporate it, as seen in Figure \ref{ex:randsent_example}.

\subsection{Quantitative Analysis} 

\paragraph{Impact of sense frequency} We examine the performance of each model as a function of the relative frequency of each sense of an ambiguous token in the training set by binning examples into frequency bins of size 20\% and present the \cf scores as a function of frequency in Table \ref{res:ambig_freq}. Low frequency senses are naturally harder to disambiguate. Most systems improve uniformly and exhibit the largest average gains in low frequency senses. While a saliency-based model is always among the strongest in a bin, \doc is never better than all of the saliency-based models.

\begin{table*}[t]
\centering
\scalebox{0.8}{
\begin{tabular}{lr|rrrrr} \toprule
    bin        & \sent & $\Delta$ \tfidf{5} & $\Delta$ \tfidf{10} & $\Delta$ \yake{5} & $\Delta$ \yake{10} &$\Delta$ \doc \\ \midrule
20-40\%  & 0.532                   & $-0.006$ & $\textbf{0.017}$ & $-0.001$ & $0.005$ & $-0.008$  \\
40-60\%  & 0.710                   & $\textbf{0.000}$ & $-0.017$ & $-0.025$ & $-0.003$ & $\textbf{0.000}$  \\
60-80\%  & 0.813                   & $-0.021$ & $-\textbf{0.011}$ & $-0.017$ & $-0.018$ & $-0.015$      \\
80-100\% & 0.847                   & $0.001$ & $0.004$ & $-0.003$ & $\textbf{0.007}$ & $0.003$       \\ \bottomrule
\end{tabular}}
\caption{\cf of various systems as a function of relative sense frequency. Best in bin \textbf{bolded}. All systems struggle with low frequency senses, but context-aware systems with $10$ salient words typically improve in this range.}
\label{res:ambig_freq}
\end{table*}

\paragraph{Impact of sentence length}

\begin{figure*}[t!]
    \centering
    \includegraphics[scale=0.65]{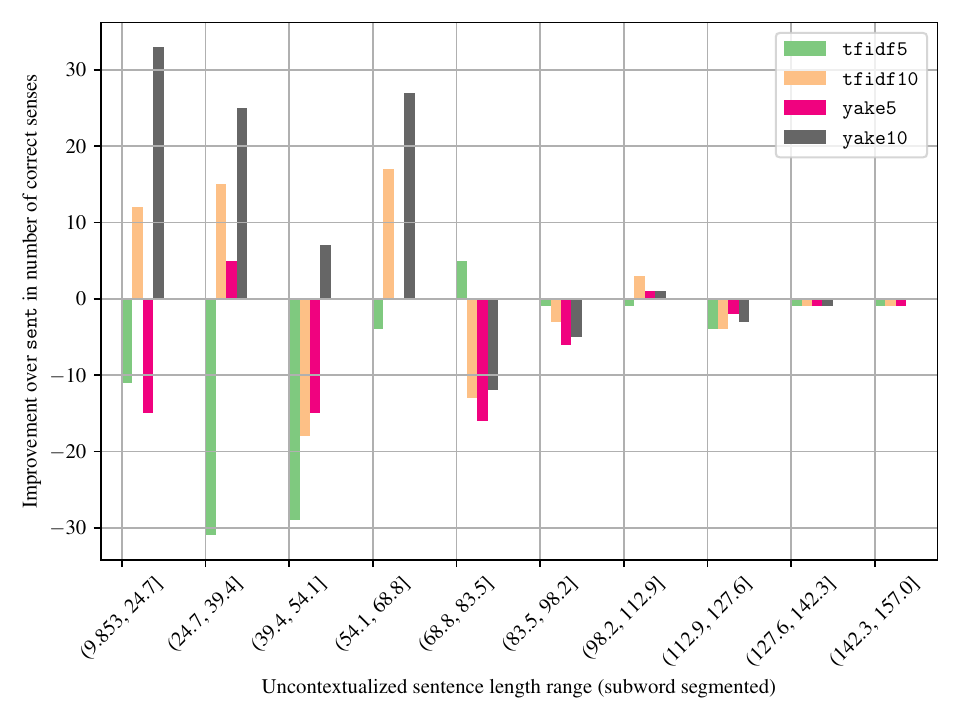}
    \caption{\# $\textsc{pos}_{\texttt{sys}}$ - \# $\textsc{pos}_{\sent}$ for each context-aware $\texttt{sys}$ as a function of subword-segmented source sentence length. Shorter sentences benefit more from context under saliency-based models with more context, with diminishing returns with more intra-sentential context.}
    \label{fig:improvement_length}
\end{figure*}

We expect that the importance of extra-sentential context diminishes with the length of $x$: as more \textit{intra-sentential} context is made available, less \textit{extra-sentential} context is required. %
As seen in Figure \ref{fig:improvement_length},  shorter sequences benefit more from context and this benefit tapers off as sequences get longer. A notable difference is that in short sentences, models with $5$ salient words are much more volatile which follows their general degraded performance relative to \sent.

\paragraph{Confidence as a function of context}

To examine the reliance on provided context, we employ conditional cross-mutual information (\textsc{CXMI}) \cite{fernandes-etal-2021-measuring} which measures the relative improvements in confidence of generating the exact reference between context-aware models and context-agnostic models over a corpus, as given by
\begin{align*}
    \operatorname{CXMI}(\hat{\textbf{W}}^k \rightarrow Y \mid X) & \approx \\ -\frac{1}{N}&\sum_{i=1}^N \log \frac{p(y_i \mid x_i)}{ p(y_i \mid x_i, \hat{\textbf{w}}^k_i)}
\end{align*}
where positive \textsc{CXMI} indicates information gain afforded by the context over the context-agnostic model and present results in Table \ref{tab:cxmi_results}. 

\begin{table}[h]
\centering
\scalebox{0.9}{
\begin{tabular}{lr} \toprule
System          & \textsc{CXMI} \\ \midrule
\tfidf{5}          & $0.002$ \\ 
\tfidf{10}         & $-0.012$ \\ 
\yake{5}          & $0.003$  \\ 
\yake{10}          & $-0.022$  \\
\doc & $-0.030$ \\ \bottomrule
\end{tabular}}
\caption{\textsc{CXMI} scores for each system show that less salient context improves model confidence, though most context-aware models are less confident than \sent.}
\label{tab:cxmi_results}
\end{table}

All models exhibit similar confidence to \sent and confidence improves with \textit{less context}. Models trained with $5$ salient words rely on their context more than both their $10$ word variants and \doc. These findings agree from previous findings that when using \textit{source} context, confidence in the translation reduces with the amount of context \cite{bawden-etal-2018-evaluating, fernandes-etal-2021-measuring} and provide some information-theoretic support for our findings. This analysis suggests that less context is more in terms of confidence. 

\paragraph{Effects of context shuffling in training}

To measure the sensitivity of context ordering imposed by our method (i.e., Footnote \ref{ordering_footnote}), we train shuffled variants of each model---$\texttt{X}^{\text{shuf}}_N$---in which the examples are prefixed with the same bag of salient words but shuffled randomly per-example prior to subword segmentation. We hypothesize that this shuffling may reduce the importance of keyword order and may improve context usage with a similar intuition to \textsc{CoWord} dropout \citep{fernandes-etal-2021-measuring}.

Using these new models, we re-run evaluation and \textsc{CXMI} analysis. We find shuffling improves the Translation Quality metrics up marginally on average, but has interesting impact on the \wsd metrics: on average, models with $5$ salient words improve in both \cprec and \crec (and subsequently \cf) while models with $10$ salient words degrade marginally. Additionally, we find interesting trends in terms of \textsc{CXMI}: confidence drops on average for systems trained with shuffled context with a slight improvement in (relatively low baseline) confidence for the \yake{10} pair of models. We refer to the full results in Tables \ref{tab:shuffled_results}--\ref{tab:shuffled_cxmi_results} in the Appendix.

\paragraph{Prefixing \sent with salient context}

 We ask whether pseudo-document context is strictly required at training time: if \sent can benefit from this context at test time with some deterministic post-processing, this suggests that the training method may not be worth the effort. We use the trained \sent system with test inputs crafted by prefixing examples with salient context from each saliency function as with saliency-based systems and treat these as inputs to \sent at test time to determine if training with salient context is necessary. For a given saliency function $\phi^{X}_N$, we label the outputs of this scheme as $\texttt{sent-X}_N$. We split the outputs on $\sep$ to mirror the expected sentence-level evaluation. Expectedly, there are steep decreases in baseline performance for \wsd Metrics and Translation Quality, suggesting that training with salient context is necessary. By examining sentence\bleu scores and manually inspecting output of low-scoring translations, we find that the large drop in translation quality can be attributed to hallucinated output and suspect that these are due to a mismatch between training and test conditions. We refer to Table \ref{tab:sentprefix_results} for full results.

Overall, saliency-based models improve translation disambiguation in more sense-frequency ranges than \doc. Saliency-based models' translation disambiguation improvements over \sent are highest for short sequences and efficacy diminishes as sentences get longer, likely due to additional intra-sentential context. All context-aware models are as confident as \sent at scoring references in \docmucow, but confidence grows as context get longer. On average, it improves over \sent's when context shuffling is introduced in training. Finally, we demonstrate that \sent cannot effectively use saliency-based context at test time, illustrating the importance of saliency-based training.

\section{Related Work}

\paragraph{Translation Disambiguation} Translating ambiguous words has been recognized as a central problem since the first formulations of \mt \cite{weaver-1952-translation} and remains an issue in state-of-the-art neural models \citep{emelin-etal-2020-detecting,campolungo-etal-2022-dibimt}. 

Much work has focused on designing explicit translation disambiguation modules for \mt, such as context-dependent translation lexicons for statistical \mt \cite[i.a.]{chan-etal-2007-word,carpuat-wu-2007-improving}.  Even in neural \mt architectures, where encoders already induce context-dependent representations of source tokens, explicitly incorporating source sentence context improves translation \citep{liu-etal-2018-handling,popescu-belis-2019-context}. We take a simpler approach by augmenting neural \mt models with salient pseudo-document context without change to standard encoder-decoder architectures.

Translation disambiguation has also been addressed by exploiting sense inventories and \wsd models to explicitly disambiguate source words for translation \citep{rios-gonzales-etal-2017-improving,nguyen-etal-2018-effect,pu-etal-2018-integrating,campolungo-etal-2022-reducing}. This direction can directly benefit from advances in \wsd \citep{bevilacqua-etal-2021-recent}, but increases the cost of training and testing \mt models in practice, due to more complex architectures and the reliance on hand-crafted resources. By contrast, our work augments \mt models with knowledge that can be easily acquired automatically at scale. 

\paragraph{Document-level \mt} Much attention has also been paid to designing \mt models that translate sentences in their document context \citep{maruf-etal-2021-survey}. This has the potential of improving many aspects of translation beyond lexical disambiguation, including pronoun translation \citep{hardmeier-2015-documentlevel}, lexical consistency \citep{carpuat-simard-2012-trouble}, and more broadly discourse cohesion and coherence \citep{jiang-etal-2022-blonde}. A wealth of models have been proposed to address this task in neural \mt, primarily by conditioning decoding on representations of preceding sentences \citep{tiedemann-scherrer-2017-neural,bawden-etal-2018-evaluating}, using cache-based mechanisms \cite{tu-etal-2018-learning,kuang-xiong-2018-fusing}, and using full document context with hierarchical attention \cite{miculicich-etal-2018-document,tan-etal-2019-hierarchical}. In this work, we show that extra-sentential context benefits \mt even when it takes the form of noisy bags of sentences rather than actual documents, and that encoding complete sentences is not necessary for \wsd.

Several avenues have been explored to compactly encode salient document context instead of directly encoding context sentences. \citet{zhang-etal-2016-topic} and \citet{wang-etal-2021-neural} use representations derived from topic models to improve \mt lexical choice. Encoding document provenance \cite{chiang-etal-2011-two} or document IDs \cite{mace-servan-2019-using} as a prefix to the input sentence can improve translation consistency, at the cost of introducing artificial tokens in the \mt vocabulary.
Here, we show that efficient general purpose keyword extraction methods benefit \wsd for \mt, but exploring the impact of other representations would be interesting to explore in the future.

\section{Conclusion}

In this work we present a simple and scalable framework for training models with salient document context to improve \wsd performance in sentence-level \mt. To test this, we release English-German \docmucow, a challenging test set for \wsd performance formed by augmenting a subset of \mucow examples with document IDs. We test our method using salient context derived from pseudo-documents formed from related sentences in ParaCrawl and show that even with noisily compiled pseudo-documents, performance surpasses that of context-agnostic sentence-level models and of comparable document-level \mt models with far more context and higher training costs. Quantitative analysis shows that shorter sentences benefit more from additional context and suggests that more context may be necessary in the case of context utilization for \wsd.

The models trained in this work use salient context constructed from pseudo-documents, in light of the relative paucity of parallel document-level contextual data.
The paradigm could, however, easily be applied to proper document-level data, were such data available.
This work focused on finding salient topical context to address lexical ambiguity, but could possibly be applied to other types of context to address other discourse phenomena. Finally, our work shows that use of pseudo-documents in training appears useful and construction of these pseudo-documents could be achieved in other ways. 

\section*{Limitations}

\paragraph{Single, high-resource setting} We demonstrate the strengths of our approach on English-German, which is a high-resource language pair. The lack of additional and diverse language pairs is a limitation. An open question is how performance changes with amount of available resources and whether additional context may improve any aspects of translation for languages with fewer resources.

\paragraph{Automatic curation}  \docmucow is constructed from an automatically curated dataset, \mucow. While inspection suggests that the resulting dataset is only minimally impacted by noise from the automation process, there are some artifacts that impact the downstream evaluation. We believe these artifacts both to be insignificant and to impact all methods equally, which we believe should not impact conclusions drawn from this work. 
Furthermore, the evaluation necessarily requires the system output to be lemmatized. This introduces two limitations: first, this methodology may not be applicable to languages without such an ability (e.g., lower resourced languages); second, this lemmatization may introduce downstream artifacts in evaluation. As before, we believe that all systems should be equally impacted by noisy lemmatizers and that this should not impact conclusions.

\paragraph{Saliency functions} While we investigate two saliency functions in this work, others could plausibly be used. We believe this to be interesting future work and present the two saliency functions in this work as a way of introducing a general framework and not as an exhaustive set.

\paragraph{Sense-enhanced models} While there has been a flurry of work examining sense-enhanced \mt, we found the resource and computational requirements to be too high to label and train models at the scale investigated in this work. We also believe that the sentence-level nature of these methods only serves to improve cases where intra-sentential context is enough to resolve the translation ambiguity while our approach provides extra-sentential context.

\paragraph{Training Data} In this work, we only use filtered ParaCrawl. While this is convenient for testing our hypothesis about the utility of \textit{pseudo}-documents and to ensure no train-test overlap, it omits a large part of high-quality bitext from available training. We ensure our baseline \sent model is reasonable by evaluating on WMT14 English-German and achieving 26.1 \bleu, which we believe is a competitive single-system score.

\paragraph{The \doc model} Because our contextualized models are effectively being trained and evaluated at the sentence-level, sampling a random sentence as context for \doc is a necessary choice. ``True'' document models require document order which is not available for the training data and, while we could plausibly train a document-level model on the small amount of curated data, it would not be a controlled experiment when compared to our models trained on ParaCrawl.

\bibliography{anthology,custom,marine}
\bibliographystyle{acl_natbib}

\newpage
\appendix

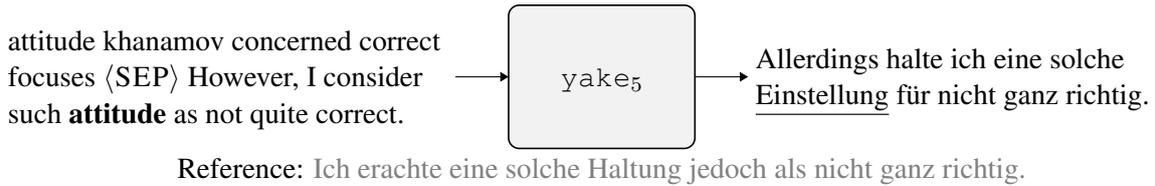
\begin{figure*}[t]
    \centering
\begin{tikzpicture}[scale=0.8]
    \matrix (m) [
        matrix of nodes,
        nodes in empty cells,
        nodes = {
            text height=2ex, text depth=0.5ex,
            inner sep=1mm, outer sep=0mm,
        },
        row sep = 0mm, column sep = 7mm,
        column 1/.style = {nodes={align=right, anchor=south east}},
        column 2/.style = {nodes={align=center, anchor=south, text width=13ex}}, %
        column 3/.style = {nodes={align=left, anchor=south west}},
    ] {
                 &                   &       \\
       \makecell[l]{attitude khanamov concerned correct \\ focuses $\sep$ However, I consider \\ such \textbf{attitude} as not quite correct.} &   \yake{5}   &   \makecell[l]{Allerdings halte ich eine solche \\ \underline{Einstellung} für nicht ganz richtig.}                \\
       &                   &                  \\
    };
    \scoped[on background layer]
        \node (enc)  [draw, rounded corners, semithick, fill=gray!10,
                      inner sep = 0mm, outer sep= 0mm,
                      fit=(m-1-2) (m-3-2)] {};
    \foreach    \i in {2}
        \draw[-{Triangle[angle=60:2pt 4]}]    (m-\i-1) -- (m-\i-2);
    \foreach    \i in {2}
        \draw[-{Triangle[angle=60:2pt 4]}]    (m-\i-2) -- (m-\i-3);
    \draw (m-3-2.south) coordinate node[below, node distance = 4em]{Reference: {\color{gray}Ich erachte eine solche \underline{Haltung} jedoch als nicht ganz richtig.}} (m.south east);
\end{tikzpicture}
    \caption{We see the context provides no information to contextualize the \textbf{ambiguous word}, so the model may not receive additional information to disambiguate.}
    \label{ex:orthogonal}
\end{figure*}

\begin{figure*}[t]
    \centering
\begin{tikzpicture}[scale=0.8]
    \matrix (m) [
        matrix of nodes,
        nodes in empty cells,
        nodes = {
            text height=2ex, text depth=0.5ex,
            inner sep=1mm, outer sep=0mm,
        },
        row sep = 0mm, column sep = 7mm,
        column 1/.style = {nodes={align=right, anchor=south east}},
        column 2/.style = {nodes={align=center, anchor=south, text width=13ex}}, %
        column 3/.style = {nodes={align=left, anchor=south west}},
    ] {
                 &                   &       \\
       \makecell[l]{good creator woman hearts stream \\ $\sep$ What good times I had with \\ my girl friends washing clothes in \\ the \textbf{stream}!} &   \tfidf{5}   &   \makecell[l]{Welche guten Zeiten hatte ich \\ mit meinen Freundinnen, die \\ Kleidung im \underline{Strom} waschen<unk>}                \\
       &                   &                  \\
    };
    \scoped[on background layer]
        \node (enc)  [draw, rounded corners, semithick, fill=gray!10,
                      inner sep = 0mm, outer sep= 0mm,
                      fit=(m-1-2) (m-3-2)] {};
    \foreach    \i in {2}
        \draw[-{Triangle[angle=60:2pt 4]}]    (m-\i-1) -- (m-\i-2);
    \foreach    \i in {2}
        \draw[-{Triangle[angle=60:2pt 4]}]    (m-\i-2) -- (m-\i-3);
    \draw (m-3-2.south) coordinate node[below, node distance = 4em]{Reference: {\color{gray}Wie kurzweilig war es, wenn wir jungen Mädchen im \textit{Bach} die Wäsche wuschen!}} (m.south east);

\end{tikzpicture}
    \caption{Some annotations in \mucow contain noise in which synonyms are assigned different sense cluster IDs which penalizes models. Here, the model produces \underline{Strom} (``stream'') instead of \textit{Bach} (``brook''), which are nearly synonymous modulo extremely narrow semantic differences. We note that the inclusion of <unk> in the output is an unfortunate effect of randomness in the subword model training.}
    \label{ex:mucow_noise}
\end{figure*}
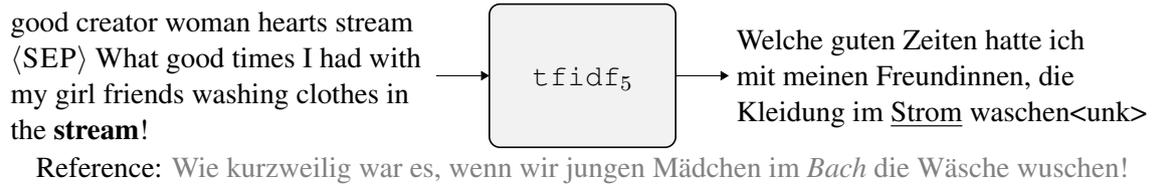

\begin{table*}[t]
\centering
\scalebox{0.9}{
\begin{tabular}{lcccrrrr}
\toprule
& \multicolumn{3}{c}{\wsd Metrics} & \multicolumn{2}{c}{Translation Quality} & \multicolumn{2}{c}{Efficiency} \\ 
System          & \cprec  & \crec & \cf & \bleu  & \comet & Length & \len \\ \midrule
\multicolumn{7}{l}{\textit{Baseline models}}\\
\sent      & 0.7850   & 0.6008   & 0.6807    & 21.8  & 0.782 & 24.3 & 97.8 ($\times 1.00$) \\ 
\doc      & 0.7830   & 0.5976   & 0.6779    & \doubleunderline{21.9}  & 0.783 & 50.5 & 135.4 ($\times 1.38$)\\ 
\multicolumn{7}{l}{\textit{Saliency-based models}}\\
\tfidf{5}          & 0.7816 &	0.5954 &	0.6759    & \doubleunderline{21.9}  & \doubleunderline{0.784} & 36.0 & 113.3 ($\times 1.16$) \\
\tfidf{10}         & \underline{0.7871} &	\underline{0.6013} &	\underline{0.6817}    & \doubleunderline{22.0}  & 0.783 & 44.6 & 127.5 ($\times 1.30$)\\ 
\yake{5}          & \underline{0.788} &	0.5973 &	0.6795     & \doubleunderline{22.0}   & \doubleunderline{0.785} & 34.3 & 110.7 ($\times 1.13$) \\
\yake{10}          & \underline{0.7885} &	\underline{0.6058} &	\underline{0.6852}     & 21.9   & 0.783 & 41.7 & 122.8 ($\times 1.26$) \\ 
\multicolumn{7}{l}{\textit{Shuffled saliency-based models}}\\
\tfidfshuf{5}    & 0.788 & \underline{0.6033} &  \underline{0.6833}   & \doubleunderline{22.0}  & \doubleunderline{0.784} & 36.0 & 115.2 ($\times 1.18$) \\
\tfidfshuf{10}         & \underline{0.7857} & 0.6002 & 0.6805   & \doubleunderline{22.0}  & 0.783 & 44.6 & 127.0 ($\times 1.30$) \\
\yakeshuf{5}          & \underline{0.7904} & \underline{0.6042} & \underline{0.6849}   & \doubleunderline{22.0}  & \doubleunderline{0.785} & 34.3 & 112.3 ($\times 1.15$) \\
\yakeshuf{10}         & 0.7860 & 0.5984 & 0.6795    & \doubleunderline{22.0}  & \doubleunderline{0.784} & 41.7 & 122.8 ($\times 1.26$) \\
\bottomrule
\end{tabular}}
\caption{Significant improvements between systems and \sent at the 95\% CI as determined by \underline{one-sided t-test} with 50 trials each with 750 samples (with replacement) and as determined by  \doubleunderline{paired bootstrap resampling} with 5k resamples. Shuffled saliency-based systems perform comparably or better than \sent and \doc in \wsd metrics and Translation Quality. Because of the size of \docmucow (20.5k sentences), large changes are unnecessary for statistical significance.}
\label{tab:shuffled_results}
\end{table*}

\begin{table*}[t]
\centering
\begin{tabular}{lrr} \toprule
 Saliency function, $\phi_N^X$          & \textsc{CXMI}, $\texttt{X}_N$ & \textsc{CXMI},  $\texttt{X}_N^{\text{shuf}}$  \\ \midrule
$\phi_{5}^{tfidf}$          & $0.002$ & $0.002$ \\ 
$\phi_{10}^{tfidf}$         & $-0.012$ & {\color{red}$-0.015$} \\ 
$\phi_{5}^{yake}$          & $0.003$ & {\color{red}$0.000$} \\ 
$\phi_{10}^{yake}$          & $-0.022$ & {\color{green}$-0.021$}  \\ \bottomrule
\end{tabular}\caption{\textsc{CXMI} scores for each system show that introducing shuffling in training {\color{red} hurts} confidence for almost all models while marginally {\color{green} improving} confidence for \yakeshuf{10}}
\label{tab:shuffled_cxmi_results}
\end{table*}

\begin{table*}[h]
\centering
\begin{tabular}{lcccrr}
\toprule
& \multicolumn{3}{c}{\wsd Metrics} & \multicolumn{2}{c}{Translation Quality} \\ 
System          & \cprec & \crec & \cf & \bleu  & \comet \\ \midrule
\multicolumn{5}{l}{\textit{Baseline model}}\\
\sent      & 0.7850   & 0.6008   & 0.6807    & 21.8  & 0.782 \\ 
\multicolumn{5}{l}{\textit{Prefixed \sent outputs}}\\
\texttt{sent-tfidf}$_5$          & 0.7755 & 0.5971 & 0.6747    & \doubleunderline{18.4}  & \doubleunderline{0.621} \\
\texttt{sent-tfidf}$_{10}$          & 0.7651 & 0.5973 & 0.6709    & \doubleunderline{15.5}  & \doubleunderline{0.513} \\
\texttt{sent-yake}$_5$          & 0.7768 & 0.5988 & 0.6763    & \doubleunderline{18.7}  & \doubleunderline{0.641} \\
\texttt{sent-yake}$_{10}$          & 0.7672 & 0.5995 & 0.6731   & \doubleunderline{15.9}  & \doubleunderline{0.535} \\ \bottomrule
\end{tabular}\caption{Performance of various prefixed outputs of \sent. Significant differences between systems and \sent at the 95\% CI as determined by \underline{one-sided t-test} with 50 trials each with 750 samples (with replacement) and as determined by  \doubleunderline{paired bootstrap resampling} with 5k resamples. Prefixed \sent models neither improve in \wsd metrics nor in translation quality illustrating the importance of training with salient context.}
\label{tab:sentprefix_results}
\end{table*}

\end{document}